\documentclass[journal,twoside]{IEEEtran}
\pdfoutput=1

\usepackage{amsmath, amssymb, amsfonts, amsthm}
\usepackage{empheq}
\usepackage{siunitx}
\theoremstyle{definition}

\usepackage[switch, pagewise]{lineno}
\usepackage{hyperref}
\usepackage{cleveref}

\crefname{equation}{eq.}{eqs.}
\crefname{figure}{fig.}{figs.}
\crefname{algorithm}{algorithm}{algorithms}

\usepackage{algorithm}
\usepackage{algorithmic}

\usepackage{subfloat}
\usepackage{graphicx}
\usepackage[caption=false, font=footnotesize]{subfig}

\usepackage{filecontents}
\usepackage{pgfplots}
\usepackage{tikzscale}
\usepackage{tikz}

\usepackage{multirow}
\usepackage{makecell}

\DeclareMathOperator*{\argmax}{arg\,max\,}
\DeclareMathOperator*{\argmin}{arg\,min\,}

\DeclareMathOperator*{\rank}{Rank}

\newcommand{\ie}{\textit{i}.\textit{e}.\ }

\title{Error Bounded Foreground and Background Modeling for Moving Object Detection in Satellite Videos}

\author{
	Junpeng Zhang, ~\IEEEmembership{Student Member, ~IEEE, }
	 Xiuping Jia, ~\IEEEmembership{Senior Member, ~IEEE } and
	 Jiankun Hu, ~\IEEEmembership{Senior Member, ~IEEE}
 }

\begin{document}
\pgfplotsset{compat=1.14}
	
\maketitle

\begin{abstract}
Detecting moving objects from ground-based videos is commonly achieved by using background subtraction techniques. 
Low-rank matrix decomposition inspires a set of state-of-the-art approaches for this task. 
It is integrated with structured sparsity regularization to achieve background subtraction in the developed method of Low-rank and Structured Sparse Decomposition (LSD).
However, when this method is applied to satellite videos where spatial resolution is poor and targets’ contrast to the background is low, its performance is limited as the data no longer fits adequately either the foreground structure or the background model.
In this paper, we handle these unexplained data explicitly and address the moving target detection from space as one of the pioneer studies. 
We propose a new technique by extending the decomposition formulation with bounded errors, named \textbf{E}xtended \textbf{L}ow-rank and \textbf{S}tructured Sparse \textbf{D}ecomposition (E-LSD). 
This formulation integrates low-rank background, structured sparse foreground as well as their residuals in a matrix decomposition problem. 
To solve this optimization problem is challenging. We provide an effective solution by introducing an alternative treatment and adopting the direct extension of Alternating Direction Method of Multipliers (ADMM). 
The proposed E-LSD was validated on two satellite videos, and experimental results demonstrate the improvement in background modeling with boosted moving object detection precision over state-of-the-art methods. 
\end{abstract}


\IEEEpeerreviewmaketitle

\section{Introduction}

\IEEEPARstart{S}{atellite}  videos, provided by JiLin-1 \cite{luo2017Jilin_1} and Skybox \cite{team2016planet}, have become available since a few years ago. 
The dense temporal information contained in these videos facilitates the solution to various surveillance problems, such as Moving Object Detection (MOD) \cite{yang2016small_vehicles_satellite_videos,xu2017rpca_vt_hsi} and target tracking \cite{mou2016tracklet_satelite_video,du2018kcf_satellite_video,zhang2018bilevel_tracking,uzkent2018tracking_hsi_cnn_kcf}.
The results will be valuable for new applications, including traffic monitoring, analysis and control at a large scale from space \cite{kopsiaftis2015vehicle_thresold}.

Detecting moving object in ground-based videos is commonly accomplished by Background Subtraction (BS) technique, after a video frame is separated into foreground and background components \cite{piccardi2004background_review,cristani2010background_subtraction_review,sobral2014cbackground_subtraction_review}.
Moving objects correspond to the regions constructed by temporally changing pixels, which is referred to the foreground, while the remaining part of the video contains those relatively stable pixels, which is commonly referred to the background.

Given that the motion of the camera is relatively small when a video is captured, the background parts between frames are temporally stable and similar, thus they are assumed to lay in a low dimensional subspace, which is commonly estimated using low-rank matrix decomposition \cite{bouwmans2014rpca_video_review, bouwmans2017DLAM, bouwmans2018rpca_formulation_bg}. 
Classical Principal Component Analysis (PCA) generates background by seeking the rank minimization of the estimated background, but it is sensitive to grossly corrupted pixels \cite{jolliffe2011PCA}. 
The moving objects take up only a tiny portion of the scene of a video so that the foreground frames are considered pixel-wise sparse. 
Robust Principal Component Analysis (RPCA) decomposes a video to a low-rank background and a sparse foreground matrix. 
Principle Component Pursuit (RPCA-PCP) \cite{wright2009RPCA_Proximal_Gradient,lin2011rpca_pcp,candes2011RPCA} and Fast Low Rank Approximation (GoDec) \cite{zhou2011GoDec} are developed to solve this low-rank matrix decomposition problem, under a deterministic condition on low rank and sparsity incoherence \cite{chandrasekaran2011rank_sparsity_incoherence}. 
In addition to the pixel-wise sparsity in RPCA, block-sparsity is also introduced to regularize the outliers in column space of the estimated background, which provides improved performance in a set of computer vision applications \cite{guyon2012rpca_block_sparsity}.
However, they are suffering from the degraded MOD performance when random noises in the estimated foreground are high, which is caused by the lack of spatial constraints on the foreground.

The first-order Markov Random Field (MRF) is then introduced to constrain smooth edges of the estimated foregrounds to reduce the effect from the random noises.
Javed \textit{et al.} \cite{javed2014orpca_mrf} shows that MRF can be utilized as a post-processing procedure to the original RPCA. 
Zhou \textit{et al.} \cite{zhou2013DECOLOR} and Shakerior \textit{et al.} \cite{shakeri2016COROLA} build unified frameworks by integrating MRF in the low-rank matrix decomposition, which works well for targets with a reasonable spatial resolution.

Another solution to integrate spatial prior in low-rank matrix decomposition is to regularize sparsity over groups of spatially neighboring pixels. 
Structure sparsity-inducing norm can measure the sparsity over groups of related variables, which is utilized to regularize structured sparsity in a various set of matrix decomposition problems \cite{jenatton2011structured_sparsity,jenatton2010sparse_hieriarchical_dictionary_learning,jia2012dictionary_learning_structured_sparsity}.
The \textbf{L}ow-rank and \textbf{S}tructured Sparse \textbf{D}ecomposition (LSD) framework \cite{liu2015LSD} utilizes the structured sparsity-inducing norm for integrating spatial prior on objects of interest to the low-rank matrix decomposition. 
While LSD decomposes a video frame into the low-rank background and the structured sparse foreground, it tends to have a large residual that does not fit these factors. 
As a result, the rank of the background is increased unnecessarily or unexpected random noises are presented in the foreground.

In satellite videos, the moving objects are often presented by only a few pixels and their contrasts to the background are low, which makes them homogeneous to random noises. 
When LSD is applied to satellite videos, the performance is then limited as it is more sensitive to the model residuals. 
In this paper, we handle these unexplained data explicitly and address the moving target detection from space as one of the pioneer studies. 
We propose a new technique by extending the decomposition formulation with bounded errors, named \textbf{E}xtended \textbf{L}ow-rank and \textbf{S}tructured Sparse \textbf{D}ecomposition (E-LSD). 
In the E-LSD, we decompose a video to the low-rank background and the structured sparse foreground, with an inequality constraint so that the residuals are bounded under a noise level. 
To solve the E-LSD formulation is challenging. We provide an effective solution by introducing an alternative treatment and adopting the direct extension of Alternating Direction Method of Multipliers (ADMM). 
Besides the potential improvements in detection performance, the E-LSD converges faster than LSD.

In summary, the main contributions of the current work are as follows.
\begin{enumerate}
	\item Satellite videoing is new, and moving object detection from space is a challenging task. 
	This is a pioneer study to address the low resolution and poor contrast data from satellites, comparing with ground-based videos.  
	
	\item We identify that the assumptions of low-rank background and the structured sparse foreground are no longer acceptable for moving object detection from space videos, as the performance is more sensitive to the model residuals.  We handle these unexplained data explicitly and developed a new error bounded background and foreground model, \textbf{E}xtended \textbf{L}ow-rank and \textbf{S}tructured Sparse \textbf{D}ecomposition (E-LSD).

	\item We provide a feasible solution to the new formulation by introducing an effective treatment and adopting the direct extension of Alternating Direction Method of Multipliers (ADMM) to meet the challenge of this hard optimization problem. 
	We demonstrate the E-LSD approach achieves improved detection performance with a reduced number of iterations using current satellite video sets, compared with the original LSD algorithm. 
	
\end{enumerate}

The remainder of this paper is organized as follows. 
The proposed formulation is detailed in \Cref{sec:method}, which is followed by the experimental evaluations in \Cref{sec:experiments}.
Finally, conclusions and suggestions for future research are given in Section \ref{sec:conclusion}.

\section{Proposed Method}
\label{sec:method}

\subsection{Structured Sparsity-Inducing Norm}

Structured sparsity-inducing norm integrates prior structures on a given set of variables. 
To encode prior structures of related variables, the entire set of variables is partitioned to a few groups by their relationships, and the structured sparsity is defined on the group levels \cite{jenatton2011structured_sparsity,yuan2006group_lasso,jenatton2011proximal_sparse_coding, xin2015gfl}.
Given a vector of $p$ variables $\mathbf{s} \in \mathbb{R}^{p}$, each group of related variables is constructed by the selected elements in $\mathbf{s}$, and the set of constructed variable groups is denoted as $\mathcal{G}$.
Structured sparsity-inducing norm sums the sparsity over all groups of variables \cite{jenatton2011structured_sparsity,jenatton2010sparse_hieriarchical_dictionary_learning} as, 
\begin{equation}
\left\Vert \mathbf{s} \right\Vert_{\ell_{1}/\ell_{\infty}} =  \sum_{g \in \mathcal{G}} \eta_{g} \max_{j \in g} \left| \mathbf{s}_{j} \right| \\
=  \sum_{g \in \mathcal{G}} \eta_{g} \left\Vert \mathbf{s}_{|g} \right\Vert_{\infty},
\end{equation}
in which $g \in \mathcal{G}$ defines a group of related variables by their indices in $\mathbf{s}$, and $\mathbf{s}_{|g} \in \mathbb{R}^{p}$ is a sparse vector with non-zero elements at the indices represented in the group $g$. 
$\eta_{g}$ assigns the weight for a group of related variables $g$. 
For simplicity, we consider each group of related variables contributes equality and set  $\eta_{g} = 1, \forall g \in \mathcal{G}$.

\begin{figure}[t]
	\footnotesize
	\centering
	\subfloat[]{\includegraphics[width=0.425 \linewidth]{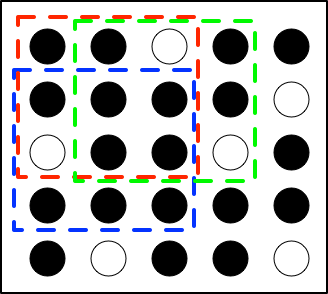}} \hfil
	\subfloat[]{\includegraphics[width=0.425 \linewidth]{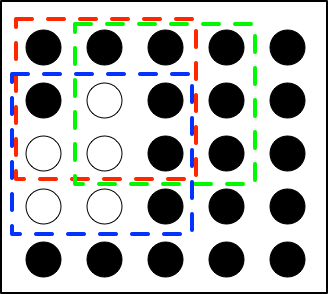}}
	\caption{An exemplar on structured sparsity. (a) and (b) are two binary images of $5 \times 5$ pixels, in which the white circles are considered as foreground pixels. The set of variable groups  $\mathcal{G}$ is constructed by a sliding grid of $3 \times 3 $, which are presented as rectangles in red, green and blue colors in for three locations, respectively. In (b), as more constructed groups contain non-zero elements, its structured sparsity-inducing norm is higher than (a). } 
	\label{fig:sparse_and_structured_sparsity}
\end{figure}

For moving object detection in satellite videos, the prior structure on the foreground is the spatial relationship among pixels.
Moving objects are commonly constructed by a blob, therefore groups of related variables $\mathcal{G}$ are constructed by spatially neighboring pixels.
Structured sparsity-inducing norm is introduced as a structured sparsity penalty to promote the sparsity at the group level, which encourages isolated random noise reduction in the foreground. 
As illustrated in \Cref{fig:sparse_and_structured_sparsity}, foreground with spatially neighboring pixels generates lower $\left\Vert \mathbf{s} \right\Vert_{\ell_{1}/\ell_{\infty}}$ than that with isolated pixels.

The construction of $\mathcal{G}$ is controlled by the prior knowledge on the moving objects. 
If $\mathcal{G}$ is defined by a set of singleton pixels, the structured sparsity-inducing norm is the conventional element-wise $\ell_{1}$ norm, which assumes pixels are independent. 
In satellite videos, the moving targets are usually very small. 
It should be proper to construct $\mathcal{G}$ via $3 \times 3$ grid scanning over the foreground.

\subsection{Extended Matrix Decomposition Model (E-LSD)}

The original \textbf{L}ow-rank and \textbf{S}tructured Sparse \textbf{D}ecomposition (LSD) \cite{liu2015LSD} defines a rank minimization problem on the background, and also imposes structured sparsity penalty on the foreground. 
Given a sequence of $n$ frames from a video noted by $\mathbf{D} \in \mathbb{R}^{p*n}$, where $p$ is the number of pixels in a frame, $\mathbf{B} \in \mathbb{R}^{p*n}$ and $\mathbf{S} \in \mathbb{R}^{p*n}$ are referred to the estimated background and foreground, respectively. 
LSD defines an optimization problem as following,
\begin{equation}
\label{eq:original_LSD}
\begin{aligned}
(\mathbf{B}^{*}, \mathbf{S}^{*}) =  \argmin_{\mathbf{B}, \mathbf{S}}&
\rank(\mathbf{B}) + 
\lambda \left\Vert \mathbf{S} \right\Vert_{\ell_{1}/\ell_{\infty}} \\ 
s.t.\ & \mathbf{D} = \mathbf{B} + \mathbf{S}
\end{aligned},
\end{equation}
in which $\lambda$ is the weight for structured sparsity-inducing norm. 
It is observed that LSD works in general for ground-based video data, although there exist unexplained data that do not fit the model. 
However, when LSD is applied to satellite videos where spatial resolution is poor and targets’ contrast to the background is low, its performance is limited as it is sensitive to the large residuals. 
In this paper, we term these unexplained data as residuals.

To explicitly handle the unexplained data in LSD, we propose an extended decomposition formulation with bounded errors, which is defined as
\begin{equation}
\label{eq:ILSD}
\begin{aligned}
(\mathbf{B}^{*}, \mathbf{S}^{*}) =  \argmin_{\mathbf{B}, \mathbf{S} }& {
	\rank(\mathbf{B}) + 
	\lambda \left\Vert \mathbf{S} \right\Vert_{\ell_{1}/\ell_{\infty}} }\\ 
s.t. & \left\Vert \mathbf{D} - \mathbf{B} - \mathbf{S} \right\Vert_{F}^{2} \le \zeta
\end{aligned},
\end{equation}
where $\zeta \ge 0$ is expected upper boundary of the unexplained data (residuals).
When $\zeta = 0$, the proposed approach degrades to the ideal case defined in the original LSD approach. 
We name this extended matrix decomposition formulation as \textbf{E}xtended \textbf{L}ow-rank and \textbf{S}tructured Sparse \textbf{D}ecomposition (E-LSD). 

As the rank minimization in \Cref{eq:ILSD} is non-convex and hard to optimize, similar to the treatment in \cite{lin2011rpca_pcp, liu2015LSD}, we relax the rank operator of $\mathbf{B}$ by its nuclear norm $\left\Vert \mathbf{B} \right\Vert_{*}$. 
The inequality constraint in \Cref{eq:ILSD} also makes the optimization problem difficult. 
One possible solution is to replace the inequality constraint by a penalty term \cite{zhou2010spcp}. 

Inspired by \cite{bouwmans2018rpca_formulation_bg, oreifej2013_3TD}, we introduce an alternative treatment which replaces the inequality constraint a decomposition model over three components, $\mathbf{D} = \mathbf{B} + \mathbf{S} + \mathbf{E}$, where the extra term $\mathbf{E} \in \mathbb{R}^{p \times n}$ is used to handle the residuals. 
\Cref{eq:ILSD} now becomes
\begin{equation}
\label{eq:relaxed_ILSD}
\begin{aligned}
(\mathbf{B}^{*}, \mathbf{S}^{*}, \mathbf{E}^{*}) =  \argmin_{\mathbf{B}, \mathbf{S}, \mathbf{E}}&{
	\left\Vert \mathbf{B} \right\Vert_{*} + 
	\lambda_{1} \left\Vert \mathbf{S} \right\Vert_{\ell_{1}/\ell_{\infty}}  + 
	\lambda_{2} \left\Vert \mathbf{E} \right\Vert_{F}^{2} }\\ 
s.t.\ & \mathbf{D} = \mathbf{B} + \mathbf{S} + \mathbf{E}
\end{aligned},
\end{equation}
where $\lambda_{1} > 0$ and $\lambda_{2} > 0$ are two scalars that assign the relative importance of  $\left\Vert \mathbf{S} \right\Vert_{\ell_{1}/\ell_{\infty}}$ and $\left\Vert \mathbf{E} \right\Vert_{F}^2$. 
In E-LSD, given a sequence of video frames $\mathbf{D}$ as input, E-LSD outputs a low-rank background matrix $\mathbf{B}$, a structured sparse foreground $\mathbf{S}$ and the unexplained residuals $\mathbf{E}$. 
When $\lambda_{2}$ tends sufficiently large with fixed $\lambda_{1}$, $\mathbf{E}$ becomes zero, then our proposed E-LSD degrades to the original LSD. 
A more detailed discussion on the selection of $\lambda_{1}$ and $\lambda_{2}$ is provided in \Cref{subsec:effect_of_e,subsec:parameter_setting}. 

\subsection{Optimization to E-LSD}

\begin{algorithm}[t]
	\caption{The Direction Extension of Alternating Direction Method of Multipliers (ADMM) for E-LSD}
	\label{alg:batch_alg}
	\begin{algorithmic}[1]
		\renewcommand{\algorithmicrequire}{\textbf{Input:}}
		\renewcommand{\algorithmicensure}{\textbf{Output:}}
		\REQUIRE $\mathbf{D} \in \mathbb{R}^{p*n}$, $\lambda_{1}>0$, $\lambda_{2} > 0$, $\mu > 1.0$, $\bar{\mu} = \mu \times \num{1.0e5}$, and $\rho > 1.0 $
		\ENSURE  $\mathbf{B}$, $\mathbf{S}$ and $\mathbf{E}$
		\STATE $\mathbf{B}_{0}= \mathbf{0}$, $\mathbf{S}_{0}=\mathbf{0}$, $\mathbf{E}_{0}=\mathbf{0}$
		\WHILE{not converged}
		\STATE Estimate $\mathbf{B}^{k+1}$ in \Cref{eq:iadm_sovler_B} using \Cref{eq:B_update}.
		\STATE Estimate $\mathbf{S}^{k+1}$ in \Cref{eq:iadm_sovler_S} using \Cref{eq:ProxFlow_dual,eq:ProxFlow_dual_to_prime}.
		\STATE $\mathbf{E}^{k+1} = \frac{\mu}{2\lambda_{2} + \mu} (\mathbf{D} - \mathbf{B}^{k+1} - \mathbf{S}^{k+1} + \frac{1}{\mu} \mathbf{Y}^{k})$.
		\STATE $\mathbf{Y}^{k+1} =  \mathbf{Y}^{k} + \mu(\mathbf{D} - \mathbf{B}^{k+1} - \mathbf{S}^{k+1} - \mathbf{E}^{k+1})$.
		\STATE $\mu = \min \{\rho \mu,  \bar{\mu} \}$, $k=k+1$.
		\STATE Check the convergence using \Cref{eq:batch_stop_criterion}
		\ENDWHILE
		\RETURN $\mathbf{B}^{k+1}$, $\mathbf{S}^{k+1}$ and $\mathbf{E}^{k+1}$
	\end{algorithmic} 
\end{algorithm}

By \Cref{eq:relaxed_ILSD}, E-LSD defines a convex optimization problem for minimizing the sum of three functions with uncoupled variables under a three-block linear constraint. 
To solve this problem, we adopt the direct extension of Alternating Direction Method of Multipliers (ADMM) \cite{chen2016direct,cai2017direct_extension_of_ADMM}, whose details are given in \Cref{alg:batch_alg}.  
We remove the equality constraint in \Cref{eq:relaxed_ILSD} by the Augmented Lagrangian Multiplier (ALM) method \cite{boyd2011ADMM}, and obtain the Lagrangian augmented optimization problem as
\begin{equation}
\label{eq:iadm_optimization}
\begin{aligned}
(\mathbf{B}^{*}, \mathbf{S}^{*}, \mathbf{E}^{*}, \mathbf{Y}^{*}) =  &\argmin_{\mathbf{B}, \mathbf{S}, \mathbf{E}, \mathbf{Y}}
\left\Vert \mathbf{B} \right\Vert_{*} + 
\lambda_{1} \left\Vert \mathbf{S} \right\Vert_{\ell_{1}/\ell_{\infty}} + 
\lambda_{2} \left\Vert \mathbf{E} \right\Vert_{F}^{2} \\   
+ & \left\langle \mathbf{Y}, \mathbf{D} - \mathbf{B} - \mathbf{S} - \mathbf{E} \right\rangle \\
+ & \frac{\mu}{2} \left\Vert  \mathbf{D} - \mathbf{B} - \mathbf{S} - \mathbf{E}  \right\Vert_{F}^{2},
\end{aligned}
\end{equation}
where $\mathbf{Y} \in \mathbb{R}^{p*n}$ is the Lagrangian multiplier and $\mu >  0$ is a positive scalar.
The augmented problem is then solved by alternatingly solving the three sub-problems as follows,
\begin{subequations}
\label{eq:iadm_sovler}
\begin{empheq}[left={\empheqlbrace\,}]{align}
\begin{split}	
\label{eq:iadm_sovler_B}
		\mathbf{B}^{k+1} = & \argmin_{\mathbf{B}} \frac{1}{\mu} \left\Vert \mathbf{B} \right\Vert_{*}   \\
		+ & \frac{1}{2} \left\Vert (\mathbf{D} - \mathbf{S}^{k} - \mathbf{E}^{k} + \frac{1}{\mu} \mathbf{Y}^{k}) - \mathbf{B} \right\Vert_{F}^{2},\\
\end{split}\\
\begin{split}
\label{eq:iadm_sovler_S}
		\mathbf{S}^{k+1} = & \argmin_{\mathbf{S}} \frac{\lambda_{1}}{\mu} \left\Vert \mathbf{S} \right\Vert_{\ell_{1}/\ell_{\infty}}  \\ 
		+ & \frac{1}{2} \left\Vert (\mathbf{D} - \mathbf{B}^{k+1} - \mathbf{E}^{k} + \frac{1}{\mu} \mathbf{Y}^{k}) - \mathbf{S} \right\Vert_{F}^{2},\\
\end{split}\\
\begin{split}
\label{eq:iadm_sovler_E}
	\mathbf{E}^{k+1} = &\frac{\mu}{2\lambda_{2} + \mu} (\mathbf{D} - \mathbf{B}^{k+1} - \mathbf{S}^{k+1} + \frac{1}{\mu} \mathbf{Y}^{k}),\\
\end{split}\\
\begin{split}
\label{eq:iadm_sovler_Y}
	\mathbf{Y}^{k+1} =  &\mathbf{Y}^{k} + \mu(\mathbf{D} - \mathbf{B}^{k+1} - \mathbf{S}^{k+1} - \mathbf{E}^{k+1}).\\
\end{split}
\end{empheq}
\end{subequations} 

In \Cref{alg:batch_alg}, sufficient conditions are satisfied to guarantee its convergence. 
$\mathbf{B}$, $\mathbf{S}$ and $\mathbf{E}$ are three separable blocks of variables. 
$\left\Vert \mathbf{B} \right\Vert_{*}$ and $\left\Vert \mathbf{S} \right\Vert_{\ell_{1}/\ell_{\infty}}$ are two convex functions, and $\left\Vert \mathbf{E} \right\Vert_{F}^{2}$ is strongly convex with modulus of 2.
When $\mu \in \{0, \frac{7\lambda_{2}}{8}\}$, it is sufficient to guarantee the convergence of our proposed algorithm, as proved in \cite{cai2017direct_extension_of_ADMM}. 
Since the theoretical convergence analysis in \cite{cai2017direct_extension_of_ADMM} is conservative and based on the worst case, in practice, empirically enlarging $\mu$ would lead to a good solution with fewer iterations.


\subsubsection{Estimation of $\mathbf{B}^{k+1}$}

At the $k+1$th iteration, $\mathbf{B}^{k+1}$ is estimated by solving the sub-problem in \Cref{eq:iadm_sovler_B} with fixed $\mathbf{S}_{k}$, $\mathbf{E}_{k}$ and $\mathbf{Y}_{k}$, whose closed-form solution is given by singular value thresholding approach \cite{wright2009RPCA_Proximal_Gradient, cai2010singular_shrinking},
\begin{equation}
\label{eq:B_update}
\begin{aligned}
\mathbf{G} = \mathbf{D} - \mathbf{S}^{k} - & \mathbf{E}^{k} + \frac{1}{\mu} \mathbf{Y}^{k}, \\
\mathbf{U} \Sigma \mathbf{V}^{T} &= \mathbf{G},\\ 
\mathbf{B}^{k+1} &= \mathbf{U} \mathcal{S}_{\frac{1}{\mu}} (\Sigma)\mathbf{V}^{T}, 
\end{aligned}
\end{equation}
where $\mathbf{U} \Sigma \mathbf{V}^{T}$ refers to the Singular Value Decomposition (SVD) of $\mathbf{G}$, and $\mathcal{S}_{\frac{1}{\mu}}(\Sigma)$ is the element-wise soft-thresholding (shrinkage) operator on the diagnose matrix $\Sigma$, which is defined as
\begin{equation}
\mathcal{S}_{\frac{1}{\mu}}(\Sigma) = \max(\Sigma-\frac{1}{\mu}, \mathbf{0}), \Sigma \succ 0, \mu > 0.
\end{equation}

\subsubsection{Estimation of $\mathbf{S}^{k+1}$}

With fixed $\mathbf{E}^{k}$, $\mathbf{Y}^{k}$ and estimated $\mathbf{B}^{k+1}$, sub-problem defined in  \Cref{eq:iadm_sovler_S} is first decomposed to a set of frame-wise optimization problems, then the solution to each of them is given by its dual problem as a Quadratic Min-cost Network Flow problem 
\footnote[1]{For simplicity, we refer $\mathbf{S}$ to $\mathbf{S}^{k+1}$ in this subsection.}.

We decompose the optimization problem in \Cref{eq:iadm_sovler_S} for a given batch to a set of frame-wise independent problems because no temporal constraints are put on the consecutive frames. 
The structured sparsity-inducing norm over the given batch can be decomposed to a combination of its frame-wise counterparts as,
\begin{equation}
\label{eq:decomposed_structure_sparsity}
\left\Vert \mathbf{S} \right\Vert_{\ell_{1}/\ell_{\infty}} = \sum_{i=1}^{n} \left\Vert \mathbf{S}_{i} \right\Vert_{\ell_{1}/\ell_{\infty}}.
\end{equation}
Let $\mathbf{H} = \mathbf{D} - \mathbf{B}^{k+1} - \mathbf{E}^{k} + \frac{1}{\mu} \mathbf{Y}^{k}$, note $\mathbf{s} = \mathbf{S}_{i}$ and $\mathbf{h} = \mathbf{H}_{i}$, where $\mathbf{S}_i$ and $\mathbf{H}_i$, $\forall i \in \{1, \cdots, n\}$, are the $i$-th column vectors in $\mathbf{S}$ and $\mathbf{H}$, respectively.
Then we rewrite the decomposed optimization problem for estimating a foreground frame as
\begin{equation}
\label{eq:ProxFlow}
\begin{aligned}
\argmin_{\mathbf{s}} &  \frac{1}{2} \left\Vert \mathbf{h} - \mathbf{s}  \right\Vert_{2}^{2} +\lambda^{\prime} \sum_{g \in \mathcal{G}} \left\Vert \mathbf{s}_{|g} \right\Vert_{\infty} 
\end{aligned},
\end{equation}
where $\lambda^{\prime} = \lambda_{1} / {\mu}$.

The unconstrained optimization problem defined in \Cref{eq:ProxFlow} cannot be solved directly by Gradient Descent method, as structured sparsity-inducing norm is non-smooth and it is challenging to derive their gradients. 
Jenatton \textit{et al.} \cite{jenatton2010sparse_hieriarchical_dictionary_learning} first show this problem can be solved by mapping to a Quadratic Min-cost Network flow problem. 
Since this structured sparsity encoding in \Cref{eq:ProxFlow} has not been explored in the field of remote sensing imaginary, we provide a brief sketch here.

To make the optimization problem defined in \Cref{eq:ProxFlow} easier to solve, auxiliary primal variables $\mathbf{z} \in \mathbb{R}^{\vert \mathcal{G} \vert }$ are introduced, where $\mathbf{z}_g$ is the corresponding variable for the group $g \in \mathcal{G}$, then the unconstrained optimization problem with inequality constraints is reformulated as,
\begin{equation}
\begin{aligned}
\argmin_{\mathbf{s}, \mathbf{z}} & \frac{1}{2} \left\Vert \mathbf{h} - \mathbf{s}  \right\Vert_{2}^{2} +\lambda^{\prime} \sum_{g \in \mathcal{G}} \mathbf{z}_g \\
s.t. & \ \forall g \in \mathcal{G}, \left\Vert \mathbf{s}_{|g} \right\Vert_{\infty} \le \mathbf{z}_g
\end{aligned},
\end{equation}
where the inequality constraint $\left\Vert \mathbf{s}_{|g} \right\Vert_{\infty} \le \mathbf{z}_g, \forall g \in \mathcal{G}$ defines the $\mathbf{z}_g$-sublevel set of $\left\Vert \mathbf{s}_g \right\Vert_{\infty}$.
Then inequality constraints are replaced by generalized conic inequalities formed by their epigraphs, as $\mathcal{C}_g = \{(\mathbf{z}_g, \mathbf{s}_{|g}) \in \mathbb{R}^{p+1} : \left\Vert \mathbf{s}_{|g} \right\Vert \le \mathbf{z}_g \}, \forall g \in \mathcal{G} $,
and the optimization problem is rewritten as
\begin{equation}
\label{eq:ProxFlow_prime}
\begin{aligned}
\argmin_{\mathbf{s}, \mathbf{z}} & \frac{1}{2} \left\Vert \mathbf{h} - \mathbf{s}  \right\Vert_{2}^{2} +\lambda^{\prime} \sum_{g \in \mathcal{G}} \mathbf{z}_g \\
s.t. & \ \forall g \in \mathcal{G}, (\mathbf{z}_g, \mathbf{s}_{|g}) \in \mathcal{C}_g
\end{aligned}.
\end{equation}

As proved in \cite{jenatton2011proximal_sparse_coding, boyd2004convex_optimization}, the convexity of the reformulated problem and the satisfied Slater's conditions \cite{slater1959slater_condition} on the generalized conic inequalities imply that strong duality holds in \Cref{eq:ProxFlow_prime}, \ie feasible solutions to primal and dual variables exist and they provide the same objective value. 
Therefore, we solve the optimization problem in \Cref{eq:ProxFlow_prime} using the primal-dual approach on its dual problem.

To remove the inequality constraints, we introduce dual variables $\tau \in \mathbb{R}^{\left| \mathcal{G} \right|}$ and $\xi \in \mathbb{R}^{p \times \left| \mathcal{G} \right|}$, and obtain the Lagrangian function of the primal problem \Cref{eq:ProxFlow_prime}
\begin{equation}
\begin{aligned}
\mathcal{L}(\mathbf{s}, \mathbf{z}, \mathbf{\tau}, \mathbf{\xi})
= \frac{1}{2} \left\Vert \mathbf{h} - \mathbf{s}  \right\Vert_{2}^{2} + \lambda^{\prime} \sum_{g \in \mathcal{G}} \mathbf{z}_g 
- \sum_{g \in \mathcal{G}}
\begin{bmatrix}
\mathbf{z}_g & \mathbf{s}_{|g}^{T}
\end{bmatrix}
\begin{bmatrix}
\tau_g \\
\xi^g
\end{bmatrix}
\end{aligned},
\end{equation}
in which $\xi^{g} \in \mathbb{R}^{p}, \forall{g} \in \mathcal{G}$ denote the corresponding dual variables for the group of variables in $g$, and $\xi$ is the set of all $\xi^{g}, \forall g \in \mathcal{G}$. 
The $\tau \in \mathbb{R}^{\vert \mathcal{G} \vert}$ is the dual variables for $\mathbf{z}$, and $\tau_{g}$ is the corresponding element for group $g \in \mathcal{G}$. 
Considering $\mathbf{s}_{|g} \in \mathbb{R}^{p}, \forall{g} \in \mathcal{G}$ is sparse, the corresponding dual variables in $\xi^{g}$ for zero elements variables in $\mathbf{s}_{|g}$ are set $\mathbf{0}$, $\xi^{g}_{j} = 0 \ \text{if} \  j \notin g$.
The dual variables $(\tau, \xi)$ also satisfy 
\footnote[2]{The inequality constraints on the dual variables are the dual of cone $\mathcal{C}_g, \forall g \in \mathcal{G}$, and the dual norm of $\ell_{\infty}$ is $\ell_{1}$ norm.}
\begin{equation}
\begin{aligned}
(\tau_{g}, \xi^{g}) \in \left\{ (\tau_{g}, \xi^{g}) \in \mathbb{R}^{p+1}: \left\Vert \xi^{g} \right\Vert_{1} \le \tau_g \right\}, 
\forall g \in \mathcal{G}.
\end{aligned}
\end{equation}

The Lagrangian dual function provides the lower bounds on the optimal value to the primal problem by seeking the minimum value of Lagrangian function over the primal variables $(\mathbf{z}, \mathbf{s})$, which is derived as
\begin{equation}
\label{eq:Lagragian_dual_function}
\begin{aligned}
&\inf_{(\mathbf{s}, \mathbf{z})} \mathcal{L}(\mathbf{s}, \mathbf{z}, \tau, \xi) = \\
&\begin{cases}
- \frac{1}{2} (\left\Vert \mathbf{h} - \sum_{g \in \mathcal{G}} \xi^{g}\right\Vert_{2}^{2} -  \left\Vert \mathbf{h} \right\Vert_{2}^{2}), 
& if \begin{cases}
\forall g \in \mathcal{G}, \lambda^{\prime} = \tau_{g} \ and \ \\
\mathbf{h} - \mathbf{s} + \sum_{g \in \mathcal{G}} \xi^{g} = 0
\end{cases} \\
-\infty  , \text{others} &
\end{cases}
\end{aligned},
\end{equation}
where all variables in $\mathbf{h}$ and $\xi$ are assumed non-negative, therefore the signs of variables $\xi$ are flipped.

Primal-dual approach searches feasible solution to primal problem by maximizing its Lagrangian dual function, and the dual problem is defined as, 
\begin{equation}
\begin{aligned}
\argmax_{\xi} \inf_{(\mathbf{s}, \mathbf{z})} \mathcal{L}(\mathbf{s}, \mathbf{z}, \tau, \xi) .
\end{aligned}
\end{equation}
As derived in \Cref{eq:Lagragian_dual_function}, the Lagrangian dual function is finite only when $\forall g \in \mathcal{G}, \lambda^{\prime} = \tau_{g}$ and $\mathbf{h} - \mathbf{s} + \sum_{g \in \mathcal{G}} \xi^{g} = 0$.
We make these equality constraints explicitly and formulate the dual problem as
\begin{equation}
\label{eq:ProxFlow_dual}
\begin{aligned}
\xi^{*} =  \argmin_{\xi} &  \frac{1}{2} \left\Vert \mathbf{h} - \sum_{g \in \mathcal{G}} \xi^{g}\right\Vert_{2}^{2} \\
s.t. \ 
& \mathbf{h} - \mathbf{s} + \sum_{g \in \mathcal{G}} \xi^{g} = 0, \\
& \forall g \in \mathcal{G},  \left\Vert \xi^{g} \right\Vert_{1} \le \lambda^{\prime} \ and \ \xi^{g}_{j} = 0 \ if \  j \notin g
\end{aligned}.
\end{equation}
The dual problem above defines a Quadratic Min-cost Network flow problem, which is defined and solved in \cite{mairal2010proxflow}.
After solving the dual problem to the decomposed optimization problem, the foreground $\mathbf{S}_i = \mathbf{s}$ is given by
\begin{equation}
\label{eq:ProxFlow_dual_to_prime}
\mathbf{s} = \mathbf{h} - \sum_{g \in \mathcal{G}}\xi^{*g},
\end{equation}
where $\xi^{*}$ is the obtained optimal solution to \Cref{eq:ProxFlow_dual}.

\subsubsection{Initialization and Termination} 
To start the E-LSD, the Lagrangian Multiplier $\mathbf{Y}_{0}$ is empirically initialized by $\mathbf{Y}_{0}= \mathbf{D} / (\left\Vert \mathbf{D} \right\Vert_{2} + \frac{1}{\lambda_{1}} \left\Vert \mathbf{D} \right\Vert_{\infty})$, as it is likely to make the objective value in \Cref{eq:iadm_optimization} reasonably large.

Similar to \cite{lin2010RPCA}, the stop criterion of the proposed algorithm is extracted by the Karush–Kuhn–Tucker (KKT) conditions \cite{boyd2004convex_optimization}
\begin{equation}
\begin{aligned}
& \mathbf{D} - \mathbf{B}^{*} - \mathbf{S}^{*} - \mathbf{E}^{*} = 0,  \\ 
\mathbf{Y} \in \partial  \left\Vert \mathbf{B}^{*} \right\Vert_{*}, 
& \mathbf{Y} \in \partial  \left\Vert \mathbf{S}^{*} \right\Vert_{\ell_{1}/\ell_{\infty}},
\mathbf{Y} \in \partial  \left\Vert \mathbf{E}^{*} \right\Vert_{2}.
\end{aligned}
\end{equation}
Due to the difficulty in computing $\partial  \left\Vert \mathbf{B}^{*} \right\Vert_{*}$ and $\partial  \left\Vert \mathbf{S}^{*} \right\Vert_{\ell_{1}/\ell_{\infty}}$, in real computing we terminate the \Cref{alg:batch_alg} by a relative stop criterion as
\begin{equation}
\label{eq:batch_stop_criterion}
\frac{\left\Vert \mathbf{D} - \mathbf{B}^{*} - \mathbf{S}^{*} - \mathbf{E}^{*}\right\Vert_{F}}{\left\Vert \mathbf{D} \right\Vert_{F}} \le \tau.
\end{equation}
The $\tau$ is influenced by the problem scale, in practice, $\tau = \num{1.0e-7}$ should be sufficient for most problems \cite{lin2010RPCA}.
\Cref{alg:batch_alg} summarizes the steps of the proposed direct extension of ADMM for E-LSD.

\section{Experiments}
\label{sec:experiments}

\begin{figure}[t]
	\footnotesize
	\centering
	\subfloat[Video 001]{\includegraphics[width=0.45 \linewidth]{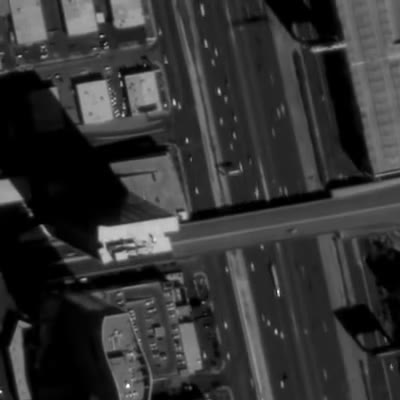}} \hfill
	\subfloat[Video 002]{\includegraphics[width=0.45 \linewidth]{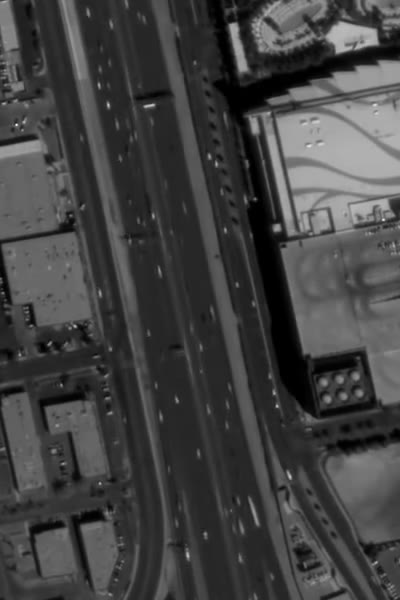}}
	
	\caption{Exemplar frames from Video 001 and Video 002.}
	\label{fig:videos}
\end{figure}

In this section, we evaluate the detection performance of E-LSD on a dataset that contains two satellite videos, namely Video 001 and Video 002. 
This dataset is constructed from a satellite video captured over Las Vagas, USA on March 25, 2014, whose spatial resolution is 1.0 meter and the frame rate is 30 frames per second. 
Both videos contained in this dataset are composed of 700 frames with boundary boxes for moving vehicles as groundtruth 
\footnote[3]{Moving vehicles are manually labeled by the Computer Vision Annotation Tool (CVAT), and a boundary box is provided for each moving object on each frame.}.
We used the first 200 frames of each video for parameter selection, and the remaining frames were used for performance evaluation. 
In this dataset, Video 001, containing complex background, is challenging for moving object detection, while the background scene is less complex in Video 002, which is mainly composed of roads, as illustrated in \Cref{fig:videos}.
More detailed information on both videos is presented in \Cref{tbl:dataset_info}.

\begin{table}
	\caption{Information on the evaluation dataset}
	\label{tbl:dataset_info}
	\centering
	\begin{tabular}{c|c|c|c|c|c}
		\hline
		\multirow{2}{*}{Video} & \multirow{2}{*}{Frame Size} & \multicolumn{2}{c|}{Cross Validation} & \multicolumn{2}{c}{Performance Evaluation} \\
		\cline{3-6}
		& & \#Frames & \#Vehicles & \#Frames & \#Vehicles \\
		
		\hline
		001 & $400 \times 400$ & 200 & 9306 & 500 & 18167 \\ 
		\hline
		002 & $600 \times 400$ & 200 & 13443 & 500 & 39362 \\ 
		\hline
	\end{tabular} 
\end{table}

\begin{figure*}
	\footnotesize
	\centering
	\begin{tabular}{c|ccc|c}
		\Xhline{1pt}
		Method	& \multicolumn{3}{c|}{LSD}	&	\textbf{E-LSD} \\
		\Xhline{1pt}
		
		Detection &
		\includegraphics[width=0.18 \linewidth]{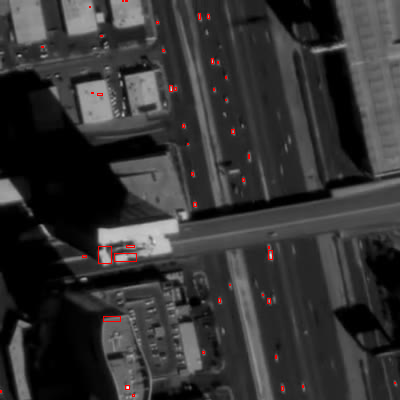}	& 
		\includegraphics[width=0.18 \linewidth]{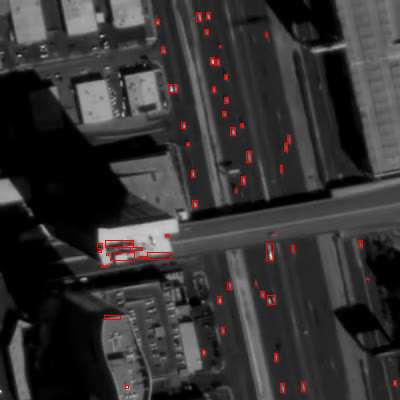}	&	
		\includegraphics[width=0.18 \linewidth]{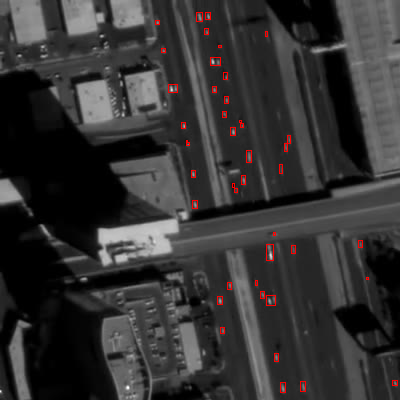}	&	
		\includegraphics[width=0.18 \linewidth]{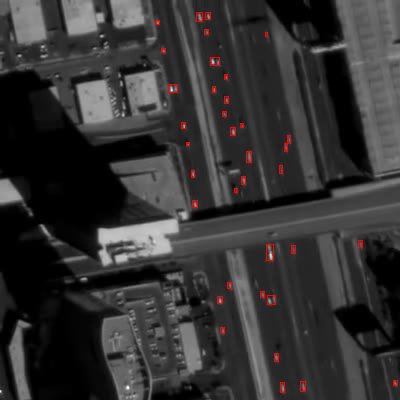}	\\
		\hline
		
		Background&
		\includegraphics[width=0.18 \linewidth]{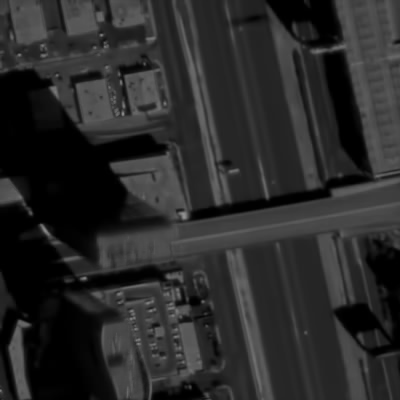}	& 
		\includegraphics[width=0.18 \linewidth]{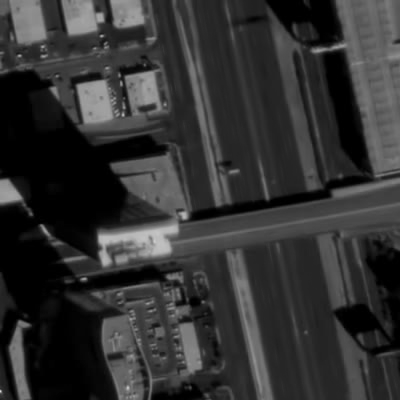}	&	
		\includegraphics[width=0.18 \linewidth]{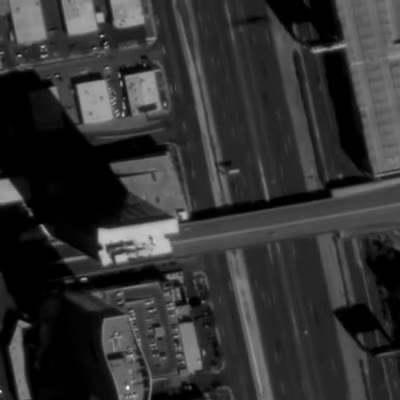}	&	
		\includegraphics[width=0.18 \linewidth]{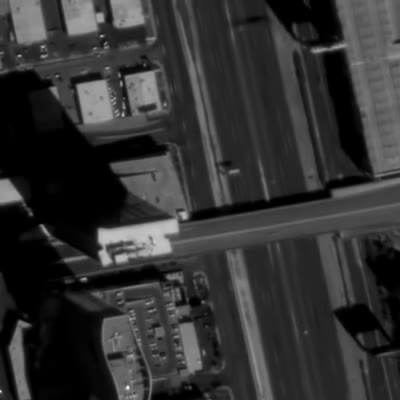} \\	
		
		\Xhline{1pt}
		
		\multirow{2}{*}{Method}	& \multicolumn{3}{c|}{LSD}	&	\multirow{2}{*}{\textbf{E-LSD}}	 \\
		\cline{2-4}
		& $\lambda=0.00025$ & $\lambda=0.00125$ & $\lambda=0.0025$ & \\
		\Xhline{1pt}
		
		Recall $\uparrow$ & 40.2\% & \textbf{90.1\%} & \underline{84.1\%} & 81.5\% \\
		\hline
		
		Precision $\uparrow$ & 36.0\% & 62.9\% & \underline{79.5\%} & \textbf{85.9\%} \\
		\hline
		
		$F_{1}$ $\uparrow$& 38.00\% & 74.85\% & \underline{81.72\%} & \textbf{83.64\%} \\
		\Xhline{1pt}
		
		$\rank(\mathbf{B})$  $\downarrow$& \textbf{1} & 89 & 200 & \underline{30} \\
		\Xhline{1pt}

	\end{tabular}
	\caption{Demonstration on the importance of introduced $\mathbf{E}$ in E-LSD.} 
	\label{fig:compare_with_lsd}
\end{figure*}

The detection performance on moving object detection is evaluated on recall, precision and $F_{1}$ scores given by
\begin{equation}
\begin{aligned}
&\text{recall} = TP/(TP + FN) \\
&\text{precision} = TP/(TP + FP) \\
&F_1 =  \frac{2 \times \text{recall} \times \text{precision}}{\text{recall} + \text{precision}}
\end{aligned},
\end{equation}
where $TP$ denotes the number of correct detections, $FN$ and $FP$ are the numbers of missed detections and false alarms, respectively.
A correct detection is defined as the Intersection over Union (IoU) against the groundtruth is greater than a threshold. 
Since the vehicles in satellite videos are small, we set the threshold as 0.3
\footnote[4]{As the estimated foreground is built by contiguous values, we utilize threshold segmentation as post-processing for extracting the foreground mask as well as the moving objects \cite{gao2012block_sparsity_bg}.}.

In addition to the evaluation of detection performance, we also consider the rank of the estimated background as an indicator of the background modeling performance. 
As the motion of the camera during a satellite video is relatively small, and the background does not change significantly, we assume better background lays in a lower dimensional subspace. 
Empirically, background with lower rank is better.

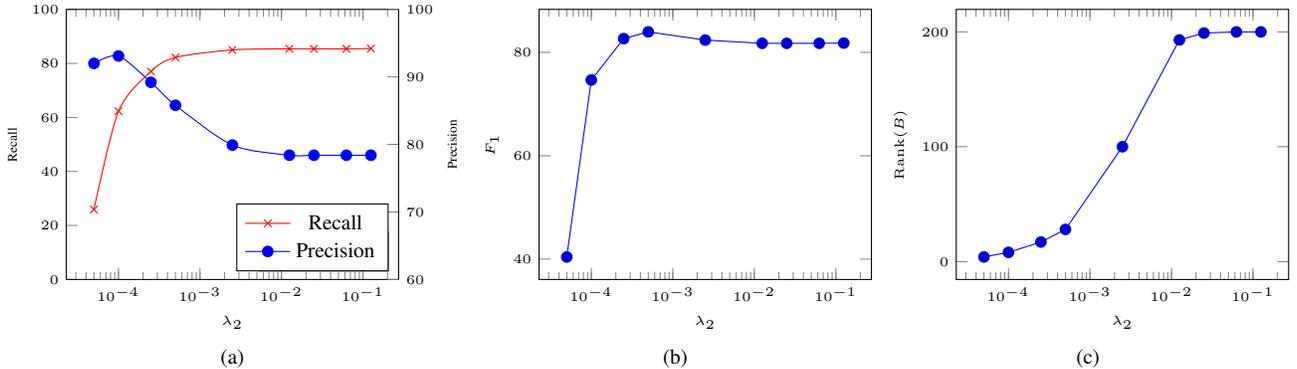
\begin{figure*}[t]
	\footnotesize
	\centering
	\subfloat[]{
		\begin{tikzpicture}
		\pgfplotsset{
			width = 6cm,
			compat=1.3,
			legend pos = south east,
		}
		\begin{axis}[
		axis y line*=left,
		ymin=0, ymax=100,
		xmode=log,
		xlabel={$\lambda_2$},
		ylabel={Recall},
		label style={font=\tiny},
		tick label style={font=\tiny}
		]
		\addplot[smooth,mark=x,red] table [x expr=0.0025 / \thisrow{factor}, y=recall, col sep=semicolon] {data/ILSD_CV_lambda_1_lambda_2_sq1.csv};\label{plot_recall}
		\addlegendentry{$recall$}
		\end{axis}
		\begin{axis}[
		axis y line*=right,
		axis x line=none,
		xmode=log,
		ymin=60, ymax=100,
		ylabel={Precision},
		xlabel near ticks,
		ylabel near ticks,
		label style={font=\tiny},
		tick label style={font=\tiny}
		]
		\addlegendimage{/pgfplots/refstyle=plot_recall}\addlegendentry{Recall}
		\addplot[smooth,mark=*,blue] table [x expr=0.0025 / \thisrow{factor}, y=precision, col sep=semicolon] {data/ILSD_CV_lambda_1_lambda_2_sq1.csv};\addlegendentry{Precision}
		\end{axis}
		\end{tikzpicture}
		\label{subfig:lambda1_lambda2_detection_sq1}
	}
	\subfloat[]{
		\begin{tikzpicture}
		\begin{axis}[ 
		width = 6cm,
		xlabel={$\lambda_2$},
		xmode=log, 
		ylabel={$F_{1}$},
		xlabel near ticks,
		ylabel near ticks,
		label style={font=\tiny},
		tick label style={font=\tiny}
		]	
		\addplot +[mark=*] table[x expr=0.0025 / \thisrow{factor}, y=f1, col sep=semicolon] {data/ILSD_CV_lambda_1_lambda_2_sq1.csv};	
		\end{axis}
		\end{tikzpicture}
		\label{subfig:F1_sq1}
	}
	\subfloat[]{
		\begin{tikzpicture}
		\begin{axis}[ 
		width = 6cm,
		xlabel={$\lambda_2$},
		xmode=log, 
		ylabel={$\rank(B)$},
		xlabel near ticks,
		ylabel near ticks,
		label style={font=\tiny},
		tick label style={font=\tiny}
		]	
		\addplot +[mark=*] table[x expr=0.0025 / \thisrow{factor}, y=rank, col sep=semicolon] {data/ILSD_CV_lambda_1_lambda_2_sq1.csv};	
		\end{axis}
		\end{tikzpicture}
		\label{subfig:rank_B_sq1}
	}
	\caption{Performance evaluation with varying $\lambda_2$ with $\lambda_{1}=0.0025$ on Video 001.} 
	\label{fig:lambda1_lambda2_sq1}
\end{figure*}

\subsection{The Evaluation on The Importance of The Introduced Bounded Error}
\label{subsec:effect_of_e}

\begin{figure*}
	
	\subfloat[$F_{1}$]{
		\begin{tikzpicture}
		\pgfplotsset{
			colormap={whitered}{color(0cm)=(white); color(1cm)=(orange!50!red)},
			legend style={at={(1.02,0.02)},anchor=north west, font= \tiny},
		}
		\begin{axis}[
		width=6.5cm,
		x dir=reverse,
		xlabel={$\lambda_{1}$},
		xmode=log,
		xmax=0.25,
		ylabel={$\lambda_{2}$},
		ymode=log,
		zmin=0.0, zmax=88.5,
		grid=major,
		view= {45}{50},
		label style={font=\tiny},
		tick label style={font=\tiny}]
		\addplot3 [surf, mesh/rows=8, shader=interp, forget plot] table[x=lambda1, y=lambda2, z expr= \thisrow{f1} * 100, col sep=comma] {data/lambda1_lambda2_grid.csv};
		
		\addplot3 [red, thick, smooth] table[x=lambda1, y=lambda2,  z expr= \thisrow{f1} * 100, col sep=comma]{data/lambda1_lambda2_grid_lambda2.csv};
		
		\addplot3 [blue, thick, smooth] table[x=lambda1, y=lambda2, z expr= \thisrow{f1} * 100, col sep=comma]{data/lambda1_lambda2_grid_lambda1.csv};
		
		\end{axis}
		\end{tikzpicture}
	}
	\subfloat[Recall]{
		\begin{tikzpicture}
		\pgfplotsset{
			colormap={whitered}{color(0cm)=(white); color(1cm)=(orange!50!red)},
			legend style={at={(1.02,0.02)},anchor=north west, font= \tiny},
		}
		\begin{axis}[
		width=6.5cm,
		x dir=reverse,
		xlabel={$\lambda_{1}$},
		xmode=log,
		xmax=0.25,
		ylabel={$\lambda_{2}$},
		ymode=log,
		zmin=0.0, 
		grid=major,
		view= {45}{50},
		label style={font=\tiny},
		tick label style={font=\tiny}
		]
		\addplot3 [surf, mesh/rows=8, shader=interp, forget plot] table[x=lambda1, y=lambda2, z expr= \thisrow{recall} * 100, col sep=comma] {data/lambda1_lambda2_grid.csv};
		
		\addplot3 [red, thick, smooth] table[x=lambda1, y=lambda2, z expr= \thisrow{recall} * 100, col sep=comma]{data/lambda1_lambda2_grid_lambda2.csv};
		
		\addplot3 [blue, thick, smooth] table[x=lambda1, y=lambda2, z expr= \thisrow{recall} * 100, col sep=comma]{data/lambda1_lambda2_grid_lambda1.csv};
		
		\end{axis}
		\end{tikzpicture}
	}
	\subfloat[Precision]{
		\begin{tikzpicture}
		\pgfplotsset{%
			colormap={whitered}{color(0cm)=(white); color(1cm)=(orange!50!red)},
			legend style={at={(0.80,1.02)},anchor=north west, font= \tiny},
		}
		\begin{axis}[
		width=6.5cm,
		x dir=reverse,
		xlabel={$\lambda_{1}$},
		xmode=log,
		xmin=0, xmax=0.25,
		ylabel={$\lambda_{2}$},
		ymode=log,
		zmin=0.0, 
		grid=major,
		view= {45}{50},
		label style={font=\tiny},
		tick label style={font=\tiny}]
		\addplot3 [surf, mesh/rows=8, shader=interp, forget plot] table[x=lambda1, y=lambda2, z expr= \thisrow{precision} * 100, col sep=comma] {data/lambda1_lambda2_grid.csv};
		
		\addplot3 [red, thick, smooth] table[x=lambda1, y=lambda2, , z expr= \thisrow{precision} * 100, col sep=comma]{data/lambda1_lambda2_grid_lambda2.csv};\addlegendentry{$\lambda_{1}=0.001$};
		
		\addplot3 [blue, thick, smooth] table[x=lambda1, y=lambda2, , z expr= \thisrow{precision} * 100, col sep=comma]{data/lambda1_lambda2_grid_lambda1.csv};\addlegendentry{$\lambda_{2}=0.005$};
		
		\end{axis}
		\end{tikzpicture}
	}
	\caption{Performance evaluation with different $\lambda_{1}$ and $\lambda_{2}$ on Video 001. The two curves in \textbf{\textcolor{red}{red}} and \textbf{\textcolor{blue}{blue}} colors are exemplars with fixed $\lambda_{1} = 0.001$ and fixed $\lambda_{2}=0.005$, respectively.}
\end{figure*}
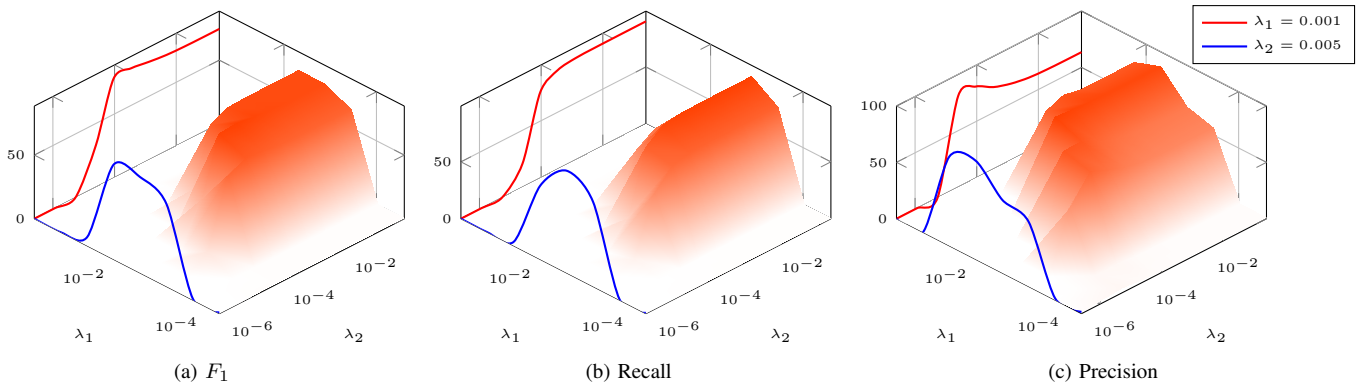

The importance of the introduced bounded errors $\mathbf{E}$ is experimentally verified, which is conducted on the Cross-Validation sequence from Video 001. 
The introduced bounded $\mathbf{E}$ helps achieve the expectations on the low-rank background and the structured sparse foreground simultaneously, which original LSD fails to achieve. 

The ignorance of the residuals $\mathbf{E}$ in LSD would result in dropped performance in either moving object detection or background modeling. 
In LSD, the parameter $\lambda$ balances the contribution of the relaxed low-rank term and structured sparsity term to the objective function. 
As $\lambda$ increases, structured sparsity is emphasized, and the residuals are more likely to be encoded in the background.
As highlighted in \Cref{fig:compare_with_lsd}, LSD achieves well detection performance when $\lambda=0.0025$, and the estimated background frame contains some residuals that do not fit to the composed model of low-rank background or structured sparse foreground.
At the same time, an increase is observed in term of the rank of the background, as illustrated in \Cref{fig:compare_with_lsd}. 
One possible reason for this increase is that some estimated background frames are so different from the others, implying that the residuals are encoded in the background model. 
On the contrary, as $\lambda$ decreases, the low-rank term contributes more to the objective function defined in \Cref{eq:original_LSD}, and the residuals tend to be encoded in the foreground frames, which hurts the detection performance. 
When $\lambda = 0.00025$, the rank of the estimated background reduces to 1, at the same time the estimated foreground is severely corrupted by the residuals, which results in the dropped detection performance. 
Therefore, without handling the residuals, the original LSD fails to achieve a good performance on both background modeling and moving object detection at the same time. 

By introducing the bounded residuals $\mathbf{E}$, our proposed E-LSD achieves both the expectations on the low-rank background and the structured sparse foreground at the same time. 
As presented in \Cref{fig:compare_with_lsd}, the extracted background of our proposed E-LSD is cleaner than that of LSD, while, at the same time, the detection performance is improved by E-LSD. 

\begin{figure}
	\begin{tikzpicture}
	\pgfplotsset{
		legend pos =north west,
	}
	\begin{axis}[
	xmin=-16, xmax=16,
	ylabel=$\lambda_{2}$,
	ymode=log, log basis y={10},
	y dir=reverse,
	xlabel=Bins,
	zlabel=Number of Pixels,
	zmin=0, 
	area plot/.style={
		fill opacity=0.75,
		draw=none,
		fill=orange!80!black,thick,
		mark=none,
		smooth
	}
	]
	
	\draw [thick, draw=red]  (axis cs: 0, 0.0001, 0) -- (axis cs: 0, 0.25, 0);
	\addplot3 [thick, blue] table [y expr= 0.0025 / \thisrowno{1}, col sep=semicolon] {data/residual_ILSD.csv};\addlegendentry{$\text{mean}(\mathbf{E})$};
	\addplot3 [area plot] table [y expr=0.0025 / \thisrowno{1}, col sep=semicolon] {data/factor_0.02_histogram_ILSD.csv};
	\addplot3 [area plot] table [y expr=0.0025 / \thisrowno{1}, col sep=semicolon] {data/factor_0.04_histogram_ILSD.csv};
	\addplot3 [area plot] table [y expr=0.0025 / \thisrowno{1}, col sep=semicolon] {data/factor_0.1_histogram_ILSD.csv};
	\addplot3 [area plot] table [y expr=0.0025 / \thisrowno{1}, col sep=semicolon] {data/factor_0.2_histogram_ILSD.csv};
	\addplot3 [area plot] table [y expr=0.0025 / \thisrowno{1}, col sep=semicolon] {data/factor_1.0_histogram_ILSD.csv};
	\addplot3 [area plot] table [y expr=0.0025 / \thisrowno{1}, col sep=semicolon] {data/factor_5.0_histogram_ILSD.csv};
	\addplot3 [area plot] table [y expr=0.0025 / \thisrowno{1}, col sep=semicolon] {data/factor_10.0_histogram_ILSD.csv};
	\addplot3 [area plot] table [y expr=0.0025 / \thisrowno{1}, col sep=semicolon] {data/factor_25.0_histogram_ILSD.csv};
	\addplot3 [area plot] table [y expr=0.0025 / \thisrowno{1}, col sep=semicolon] {data/factor_50.0_histogram_ILSD.csv};
	\end{axis}
	\end{tikzpicture}
	\caption{Histogram of $\mathbf{E}$ with varying $\lambda_2$ with fixed $\lambda_{1}=0.0025$ on Video 001}
	\label{fig:residual_histo}
\end{figure}
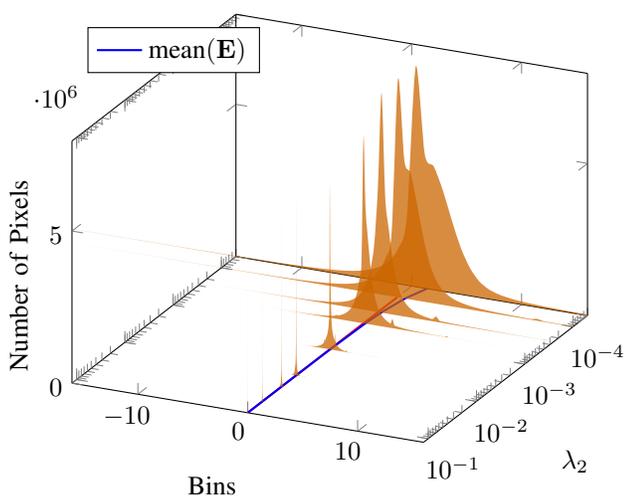

To further investigate the effect of $\mathbf{E}$, we vary its weight $\lambda_{2}$ with fixed $\lambda_{1} = 0.0025$ in E-LSD defined in \Cref{eq:relaxed_ILSD}.
When $\lambda_{2}$ is sufficiently large, $\mathbf{E}$ tends zero, then the E-LSD degrades to LSD. 
As $\lambda_{2}$ decreases, the residual term is less emphasized, and the upper bound for $\mathbf{E}$ starts to increase, which allows residuals to be encoded by the introduced $\mathbf{E}$.
As illustrated in \Cref{subfig:rank_B_sq1}, with decreasing $\lambda_{2}$, the rank of the estimated background decreases dramatically. 
At the same time, the precision of detected moving objects increases, as shown in \Cref{subfig:lambda1_lambda2_detection_sq1}, which may be attributed to the removal of possible residuals from the foreground by $\mathbf{E}$. 
When $\lambda_2$ becomes extremely tiny, the upper bound for $\mathbf{E}$ turns very large, which may include too much information into $\mathbf{E}$ and break the detection performance. 
As shown in \Cref{fig:residual_histo}, the shifts of the mean values of $\mathbf{E}$ and the expansion of the corresponding histograms demonstrate that the more pixel entries are included by the residual term as $\lambda_{2}$ decreases, which explains the drop of recall scores on the other hand. 
Similar trend of the detection performance with different $\lambda_{2}$ can also be observed with different selection on $\lambda_{1}$, as presented in \Cref{fig:lambda1_lambda2_sq1}. 

\begin{figure*}[t]
	\footnotesize
	\centering
	\includegraphics[width=0.985 \linewidth]{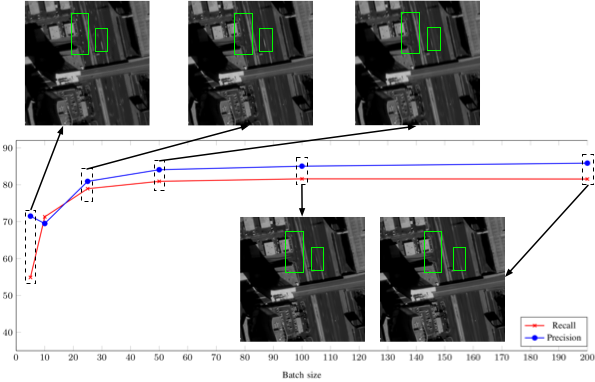}
	\caption{Performance evaluation on batches of different lengths. Exemplar estimated backgrounds are provided at increasing batch lengths, where two sub-regions are highlighted for containing slow-moving targets and large targets, respectively} 
	\label{fig:batch_size}
\end{figure*}

\begin{table*}
	\caption{Detection Performance Evaluation.}
	\label{tbl:detection_results}
	\centering
	\begin{tabular}{c|cccc|cccc|c}		
		\hline 
		\multirow{2}{*}{Video} & \multicolumn{4}{c|}{001} & \multicolumn{4}{c|}{002} & \multirow{2}{*}{Avg($F_1$) $\uparrow$}\\
		\cline{2-9}
		& Recall $\uparrow$ &  Precision $\uparrow$ &  $F_{1}$ score $\uparrow$ & $\rank(\mathbf{B})\downarrow$ 
		& Recall $\uparrow$ &  Precision $\uparrow$ &  $F_{1}$ score $\uparrow$	& $\rank(\mathbf{B})\downarrow$ & \\ 
		\hline 
		RPCA-PCP 	& \underline{94.4\%} &40.8\% & 56.98\% & 2 & \textbf{90.2\%} &78.1\% & 83.72\% & 2 & 70.35\% \\ 
		\hline 
		GoDec		 	& \textbf{94.7\%} & 35.7\%  & 51.85\% & 6 & \underline{89.9\%} & 80.9\%  & 85.16\% & 6 & 68.51\% \\ 
		\hline 
		DECOLOR 	& 77.4\% & 58.6\%  & 66.70\% & 2 & 79.1\% & 80.8\%  & 79.55\% & 2 & 73.13\%\\ 
		\hline
		LSD	& 86.8\% & \underline{70.8\%}  & \underline{77.98\%} & 497 & 82.2\% & \underline{90.9\%}  & \textbf{86.31\%} & 496 &  \underline{82.15\%} \\
		\hline 
		\textbf{E-LSD} & 85.0\% & \textbf{78.9\%}  &  \textbf{81.82\%} & 65 & 79.6\% & \textbf{93.8\%}  &  \underline{86.13\%} & 76 & \textbf{83.98\%} \\
		\hline
	\end{tabular} 
\end{table*}

\subsection{Parameter Setting}
\label{subsec:parameter_setting}

The E-LSD is mainly controlled by the weights of sparsity term and the residual term, $\lambda_1$ and $\lambda_2$. 
As varying $\lambda_{2}$ with fixed $\lambda_{1}$ has been discussed in \Cref{subsec:effect_of_e}, we mainly focus on the effects of different $\lambda_{1}$ on the detection performance. 

The parameter $\lambda_{1}$ controls the importance of the structured sparsity term in the objective function of E-LSD. 
As $\lambda_{1}$ increases from an extremely small number, structured sparsity term starts contributing more to the objective function, and the foreground is expected to be more structured sparse, which improves the detection performance.
If $\lambda_{1}$ turns too large, the structured sparsity term tends to prevent moving objects from being encoded in the foreground, which causes a drop in the detection performance. 
As presented in \Cref{fig:lambda1_lambda2_sq1}, with fixed $\lambda_{2}$, the detection performance by E-LSD increases, when $\lambda_{1}$ gradually increases from $10^{-4}$. 
After achieving the best performance of E-LSD, the detection performance starts decreasing as $\lambda_{1}$ continues increasing.

In addition to the trade-off between $\lambda_1$ and $\lambda_2$, the performance of E-LSD is also influenced by the length of the batch for processing, and a considerable batch length would benefit both detection and background estimation performance.
Estimation of background entries for slow moving objects or large objects requires more extended observation sequence, otherwise, the backgrounds would contain blurred foreground entries, leading to a dropped detection performance. 
As presented in \Cref{fig:batch_size},  we gradually increase the length of the batch from 5 to 200, and this increase results in improved recall and prevision.
At the same time, the two highlighted regions in \Cref{fig:batch_size}, which are corresponding to containing slow moving objects and large objects, respectively, are displayed as examples for background estimation. We can see the backgrounds detected are more homogeneous, as the length of batch increases.

In following experiments we present in this paper, $\lambda_{1} = 1/ {\sqrt{p}}$ and $\lambda_{2}= \lambda_{1} / 5$ were set for E-LSD, and the corresponding batch size in E-LSD was selected as the length of the video for processing. 
Further fine-tuning on the selection of $\lambda_{1}$ and $\lambda_{2}$ may improve the performance more.

\begin{figure*}[t]
	\footnotesize
	\centering
	\begin{tabular}{ccccc}
		\centering

		\includegraphics[width=0.18 \linewidth]{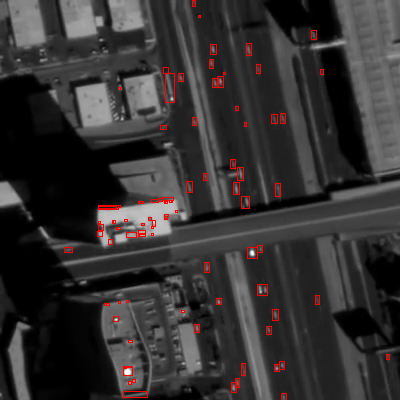}	&	
		\includegraphics[width=0.18 \linewidth]{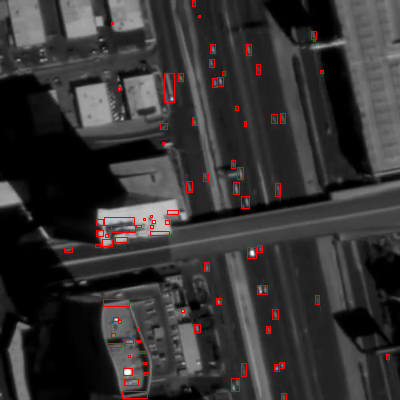}	&	
		\includegraphics[width=0.18 \linewidth]{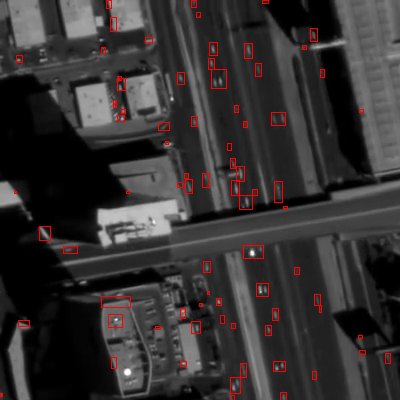}	& 
		\includegraphics[width=0.18 \linewidth]{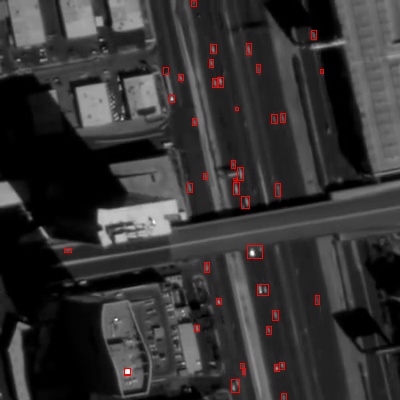}	&	
		\includegraphics[width=0.18 \linewidth]{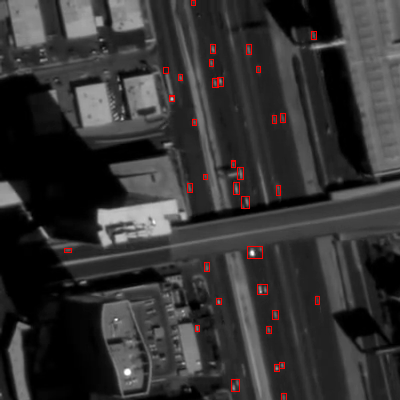} \\
		
		\includegraphics[width=0.18 \linewidth]{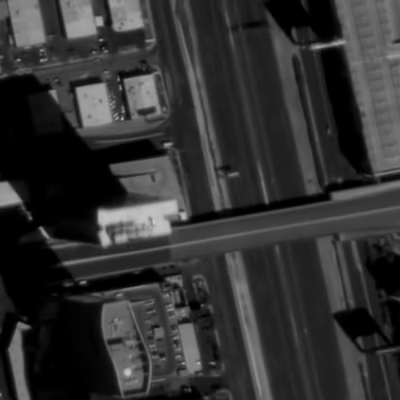}	&	
		\includegraphics[width=0.18 \linewidth]{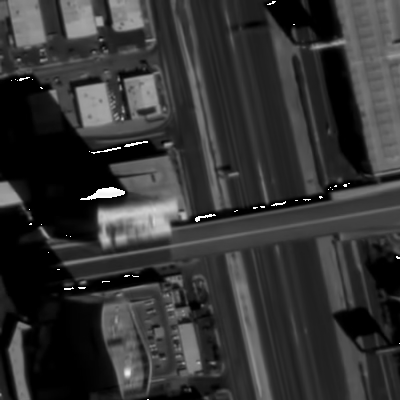}	&	
		\includegraphics[width=0.18 \linewidth]{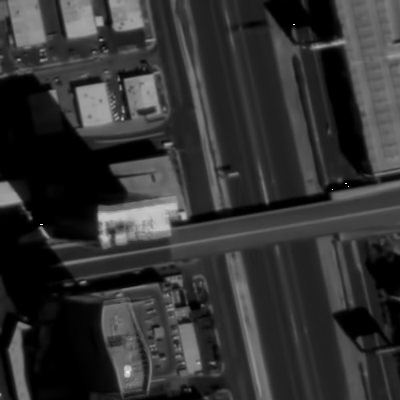}	& 
		\includegraphics[width=0.18 \linewidth]{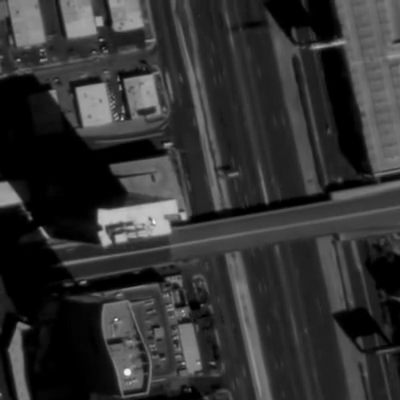}	&	
		\includegraphics[width=0.18 \linewidth]{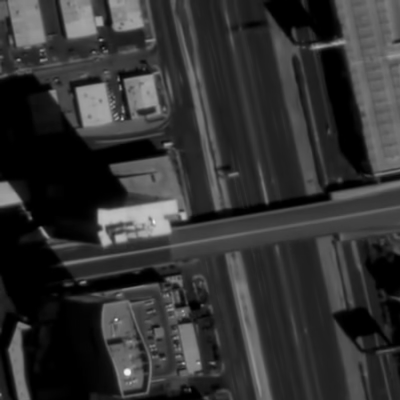} \\
		
		\includegraphics[width=0.18 \linewidth]{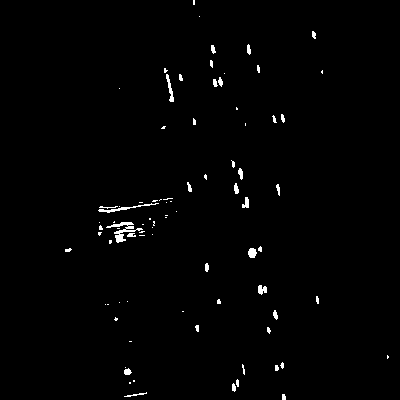}	&	
		\includegraphics[width=0.18 \linewidth]{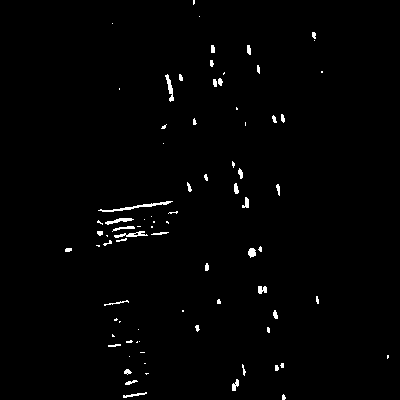}	&	
		\includegraphics[width=0.18 \linewidth]{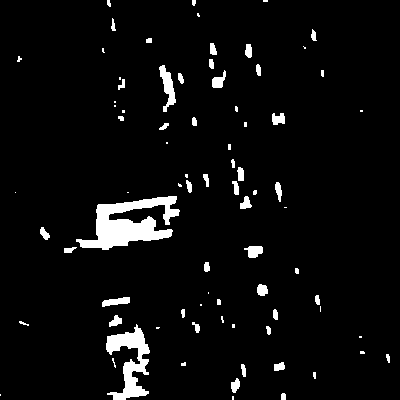}	& 
		\includegraphics[width=0.18 \linewidth]{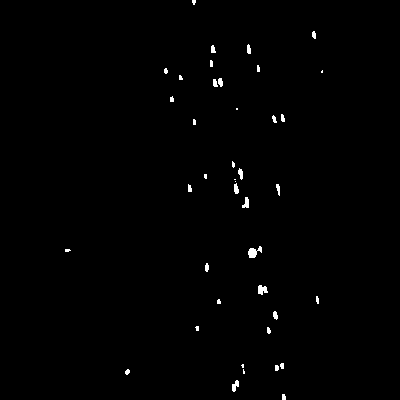}	&	
		\includegraphics[width=0.18 \linewidth]{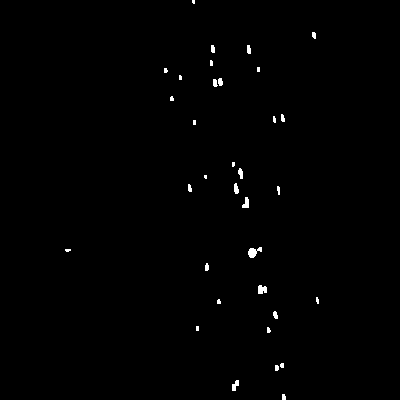} \\
		
		RPCA-PCP	&	GeDec	&	DECOLOR	&	LSD	&	\textbf{E-LSD}	 \\
	\end{tabular}
	\caption{Comparison with other existing state-of-the-art methods. Each column presents different approaches. The first row are the visualized detection results, which are all the $250^{th}$ frame, and the second row and third row are the estimated backgrounds and foreground marks.} 
	\label{fig:detection_visualization}
\end{figure*}

\subsection{Performance Evaluations}

To verify the effectiveness of E-LSD, we compare the detection performance against 4 state-of-the-art approaches, which are RPCA-PCP \cite{candes2011RPCA}, Godec \cite{zhou2011GoDec}, DECOLOR \cite{zhou2013DECOLOR} and the original LSD \cite{liu2015LSD}. 
RPCA-PCP and Godec are low rank matrix decomposition methods with element-wise sparsity on the foreground, which are solved by Principal Component Pursuit (PCP) and Fast Low Rank Approximation, respectively. 
DECOLOR imposes contiguous outlier constraints on low rank decomposition by introducing the first-order Markov Random Filed.

As presented in \Cref{tbl:detection_results}, the E-LSD algorithm achieves the highest precision on both videos and comparable performance in term of recall. 
Video 001 contains complex background, which makes it challenging for moving object detection. 
Compared with the RPCA-PCP and GoDec algorithms, E-LSD boosts the detection precision by the introduced structure sparsity. 
Secondly, we can see LSD and E-LSD can overcome the shortcoming of adding targets’ neighboring background pixels to the foreground by DECOLOR, as presented in \Cref{fig:detection_visualization}. 
Compared with the original LSD approach, the E-LSD significantly reduces the rank of the estimated background from 496 to 76, which means the background contains less undetected moving objects. 
As demonstrated in \Cref{fig:detection_visualization}, fewer residuals are included in the estimated foreground by E-LSD than LSD.
The other video in this dataset, Video 002, is less challenging, most algorithms achieve good performance on it, and the improvement of E-LSD is less significant. 
E-LSD and LSD achieve similar detection performances on Video 002, and the estimated background by E-LSD has a lower rank than LSD, implying that E-LSD improves the background modeling performance.
Overall, the E-LSD significantly outperforms RPCA-PCP, GoDec, and DECOLOR. 
Compared with the LSD, the E-LSD improves the performance on both moving object detection and background modeling.

\begin{figure}[t]
	\centering
	\subfloat[Video 001]{
		\begin{tikzpicture}
		\pgfplotsset{
			width = 9 cm,
			height = 5 cm,
			legend style={at={(0.02,0.02)}, font=\tiny, anchor=south west},
		}
		\begin{axis}[
		xlabel={IoU Threshold},
		ylabel={$F_{1}$},
		xmin=0.05, xmax=0.95,
		xlabel near ticks,
		ylabel near ticks,
		label style={font=\tiny},
		tick label style={font=\tiny}
		]	
		\addplot +[thick, red, dashed, dash pattern=on 3pt off 3pt, mark=none] table[x=iou, y expr=\thisrow{recall} * \thisrow{precision} * 2/(\thisrow{recall} +\thisrow{precision})*100, col sep=semicolon] {data/recall_precision_iou_sq_1/RPCA_PCP.csv};\addlegendentry{RPCA-PCP}
		\addplot +[thick, blue, dashed, dash pattern=on 4pt off 2pt on 1pt off 2pt, dash phase=4pt, mark=none] table[x=iou, y expr=\thisrow{recall} * \thisrow{precision}* 2/(\thisrow{recall} +\thisrow{precision})*100, col sep=semicolon] {data/recall_precision_iou_sq_1/GoDec.csv};\addlegendentry{GoDec}
		\addplot +[thick, blue, mark=none] table[x=iou, y expr=\thisrow{recall} * \thisrow{precision}* 2/(\thisrow{recall} +\thisrow{precision})*100, col sep=semicolon] {data/recall_precision_iou_sq_1/DECOLOR.csv};\addlegendentry{DECOLOR}
		\addplot +[thick, black, mark=none] table[x=iou, y expr=\thisrow{recall} * \thisrow{precision}* 2/(\thisrow{recall} +\thisrow{precision})*100, col sep=semicolon] {data/recall_precision_iou_sq_1/LSD.csv};\addlegendentry{LSD}
		\addplot +[thick, red, mark=none] table[x=iou, y expr=\thisrow{recall} * \thisrow{precision}* 2/(\thisrow{recall} +\thisrow{precision})*100, col sep=semicolon] {data/recall_precision_iou_sq_1/ILSD.csv};\addlegendentry{\textbf{E-LSD}}
		\end{axis}
		\end{tikzpicture}
	}\\
	\subfloat[Video 002]{
		\begin{tikzpicture}
		\pgfplotsset{
			width = 9 cm,
			height = 5 cm,
			legend style={at={(0.02,0.02)}, font=\tiny, anchor=south west},
		}
		\begin{axis}[
		xlabel={IoU Threshold},
		ylabel={$F_{1}$},
		xmin=0.05, xmax=0.95,
		xlabel near ticks,
		ylabel near ticks,
		label style={font=\tiny},
		tick label style={font=\tiny}
		]	
		\addplot +[thick, red, dashed, dash pattern=on 3pt off 3pt, mark=none] table[x=iou, y expr=\thisrow{recall} * \thisrow{precision}* 2/(\thisrow{recall} +\thisrow{precision})*100, col sep=semicolon] {data/recall_precision_iou_sq_2/RPCA_PCP.csv};\addlegendentry{RPCA-PCP}
		\addplot +[thick, blue, dashed, dash pattern=on 4pt off 2pt on 1pt off 2pt, dash phase=4pt, mark=none] table[x=iou, y expr=\thisrow{recall} * \thisrow{precision}* 2/(\thisrow{recall} +\thisrow{precision})*100, col sep=semicolon] {data/recall_precision_iou_sq_2/GoDec.csv};\addlegendentry{GoDec}
		\addplot +[thick, blue, mark=none] table[x=iou, y expr=\thisrow{recall} * \thisrow{precision}* 2/(\thisrow{recall} +\thisrow{precision})*100, col sep=semicolon] {data/recall_precision_iou_sq_2/DECOLOR.csv};\addlegendentry{DECOLOR}
		\addplot +[thick, black, mark=none] table[x=iou, y expr=\thisrow{recall} * \thisrow{precision}* 2/(\thisrow{recall} +\thisrow{precision})*100, col sep=semicolon] {data/recall_precision_iou_sq_2/LSD.csv};\addlegendentry{LSD}
		\addplot +[thick, red, mark=none] table[x=iou, y expr=\thisrow{recall} * \thisrow{precision}* 2/(\thisrow{recall} +\thisrow{precision})*100, col sep=semicolon] {data/recall_precision_iou_sq_2/ILSD.csv};\addlegendentry{\textbf{E-LSD}}
		\end{axis}
		\end{tikzpicture}
	}
	\caption{Detection performance over varying IoU thresholds.}
	\label{fig:f1_iou_datasets1}
\end{figure}
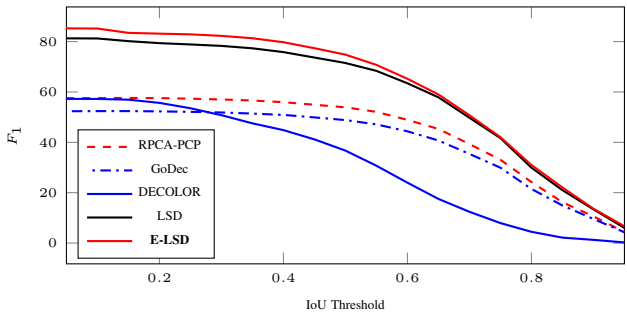
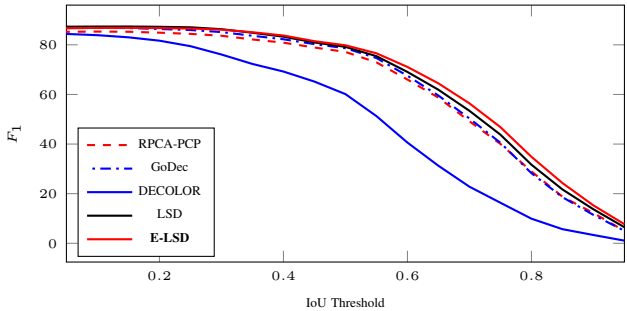

We also evaluate the detection performance with varying IoU thresholds. 
As illustrated in \Cref{fig:f1_iou_datasets1}, on Video 001, the $F_{1}$ score deceases as the IoU threshold increases. 
Compared with RPCA-PCP, GoDec, and DECOLOR, the E-LSD significantly improves the detection performance at all thresholds, and slight improvement against LSD is also observed. 
Since Video 002 is less challenging, all methods except DECOLOR perform similarly.
The E-LSD achieves higher $F_{1}$ score when the threshold is large. 

\subsection{Analysis on Computation Consumption}

\begin{figure}
	\centering		
	\begin{tikzpicture}
	\pgfplotsset{
		legend style={font=\tiny},
		legend pos= north east,
	}
	\begin{axis}[
		width = 9 cm,
		height = 5 cm,
		xlabel={Iteration},
		ylabel={Stop Criterion},
		xmin=0.0, xmax=60,
		ymode=log, log basis y={10},
		xlabel near ticks,
		ylabel near ticks,
		label style={font=\tiny},
		tick label style={font=\tiny},
		]	
		
	\addplot +[thick, red, smooth, mark=none] table[x=iter, y = stops, col sep=semicolon] {data/convergence/ILSD_sq_1.csv};\addlegendentry{E-LSD (Video 001)}
	\addplot +[thick, black, smooth, mark=none] table[x=iter, y = stops, col sep=semicolon] {data/convergence/LSD_sq_1.csv};\addlegendentry{LSD (Video 001)}

	\addplot +[thick, red, dashed, mark=none] table[x=iter, y=stops, col sep=semicolon] {data/convergence/ILSD_sq_2.csv};\addlegendentry{E-LSD (Video 002)}
	\addplot +[thick, black, dashed, mark=none] table[x=iter, y=stops, col sep=semicolon] {data/convergence/LSD_sq_2.csv};\addlegendentry{LSD (Video 002)}
	\end{axis}
	\end{tikzpicture}
	\caption{Comparison on convergence between E-LSD and LSD approach.}
	\label{fig:convergence}
\end{figure}
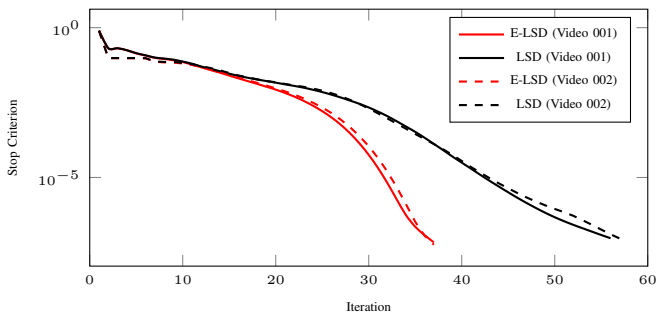

In addition to the improved detection performance, another advantage of the E-LSD is its faster convergence compared with the original LSD approach. 
In the LSD, $\mathbf{E}$ factor is ignored, which has been implicitly factored to either the foreground or the background, which challenges the optimization procedure, and results in slower convergence. 
As illustrated in \Cref{fig:convergence}, as the relative stop criterion becomes small, the E-LSD reaches the same level of relative stop criterion earlier than LSD. 
By using the same relative stop criterion, the E-LSD requires about 20 fewer iterations, as demonstrated in \Cref{tbl:num_iteration}.
This improvement should be owned to the explicit handling of the extra data structure beyond the low-rank background and the sparse foreground. 

\begin{table}
	\caption{Comparison on Number of Iterations with Relative Stop Criterion}
	\label{tbl:num_iteration}
	\centering
	\begin{tabular}{c|c|c|c|c}
		\hline
		\multirow{2}{*}{Video}  & \multicolumn{2}{c|}{Video 001} & \multicolumn{2}{c}{Video 002} \\
		\cline{2-5}
		& \# Iteration & Time (s) & \# Iteration & Time (s) \\
		
		\hline
		LSD &  56 & 34238.43 & 57 & 59748.96 \\ 
		\hline
		\textbf{E-LSD} &  \textbf{37} & \textbf{11938.70} & \textbf{37} & \textbf{23038.05} \\ 
		\hline
	\end{tabular} 
\end{table}

In terms of the computation consumption, the E-LSD significantly reduces the time cost in the process, as fewer iterations are required. 
In the LSD and the E-LSD, Singular Value Decomposition (SVD) and structured sparsity encoding are involved in each optimization iteration, and they take the majority of the time cost. 
Reducing the iterations in the E-LSD means SVD and structured sparsity encoding.
Although additional computation is introduced in the E-LSD for updating $\mathbf{E}$, these operations are basic matrix sum operators, whose cost is negligible.
As demonstrated in \Cref{tbl:num_iteration}, the time for processing a video with same relative stop criterion, $\tau = 1.0e^{-7}$, is significantly reduced by the E-LSD \footnote[5]{The time costs are measured on a laptop with Intel Core i7 4-core CPU (2.2 GHz) and 16 GB RAMs, and both algorithms are implemented by Python and C++ with the same level of optimization.}.
Therefore, the E-LSD is more cost-efficient, compared with the LSD.

\section{Conclusion}
\label{sec:conclusion}

Moving Object Detection (MOD) aims to reveal moving objects from a video or a sequence of images, after which the extracted objects can be associated for object tracking.
Therefore, MOD in satellite videos is a fundamental and critical task in satellite video processing.  
Detecting moving objects is commonly achieved by background subtraction techniques.
Low rank and structured sparse matrix decomposition integrates the prior structure of foreground with spatial information to background subtraction, however, it often ends up with high model residuals. 

In order to handle these ignored data, we propose an \textbf{E}xtended \textbf{L}ow-rank and \textbf{S}tructured Sparse \textbf{D}ecomposition (E-LSD), where those outliers are bounded under a noise level. 
Experimental results show that the proposed E-LSD achieves boosted detection precision as well as improved $F_{1}$ scores.
In addition to the improved detection performance, by handling the outliers explicitly the E-LSD converges faster than the original LSD approach.
As more satellite video data are available, more extensive testing can be conducted in the future study. 

\section*{Acknowledgment}

This work is partially supported by China Scholarship Council. The authors would like to thank Planet Team for providing the data in this research \cite{team2016planet}.

\bibliographystyle{IEEEtran}
\footnotesize\bibliography{refs}

\begin{IEEEbiography}[{\includegraphics[width=0.95 \textwidth]{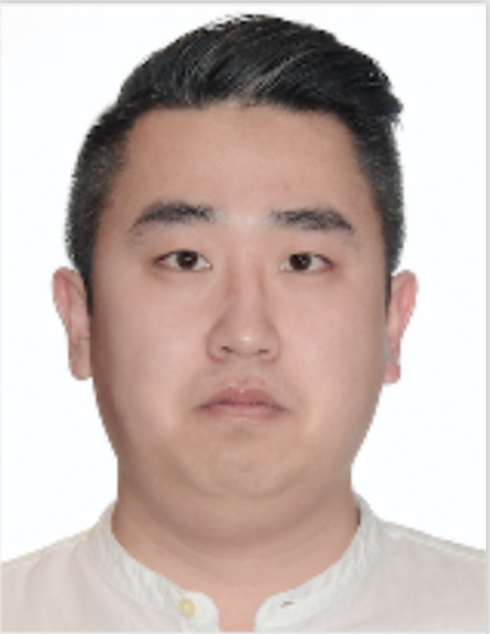}}]{Junpeng Zhang}
	received the B.Sci. degree from the China University of Mining and Technology, Xuzhou, China, in 2013 and the Master's degree in surveying engineering from the same university in 2016. 
	He is currently pursuing the Ph.D. degree in electrical engineering from The University of New South Wales, Australia.
	His research interests include object detection and tracking in remote sensing imaginary.
	He was the winner of "DSTG Best Contribution to Science Award" in Digital Image Computing: Techniques and Applications 2018 (DICTA 2018). 
	
\end{IEEEbiography}

\begin{IEEEbiography}[{\includegraphics[width=0.95 \textwidth]{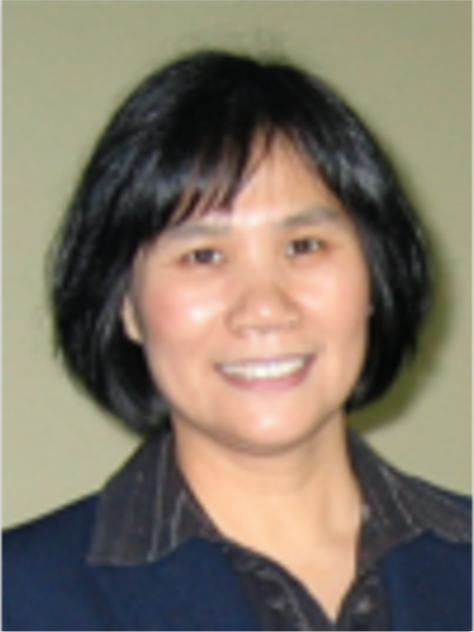}}]{Xiuping Jia}
	(M’93–SM’03) received the B.Eng. degree from the Beijing University of Posts and Telecommunications, Beijing, China, in 1982 and the Ph.D. degree in electrical engineering from The University of New South Wales, Australia, in 1996. Since 1988, she has been with the School of Engineering and Information Technology, The University of New South Wales at Canberra, Australia, where she is currently an Associate Professor. 
	Her research interests include remote sensing, image processing, and spatial data analysis. Dr. Jia has authored or coauthored more than 200 referred papers, including over 100 journal papers with h-index of 34 and i10 of 102. 
	She has co-authored of the remote sensing textbook titled Remote Sensing Digital Image Analysis [Springer-Verlag, 3rd (1999) and 4th eds. (2006)]. 
	She is a Subject Editor for the Journal of Soils and Sediments and an Associate Editor of the IEEE TRANSACTIONS ON GEOSCIENCE AND REMOTE SENSING.
	
\end{IEEEbiography}

\begin{IEEEbiography}[{\includegraphics[width=0.95 \textwidth]{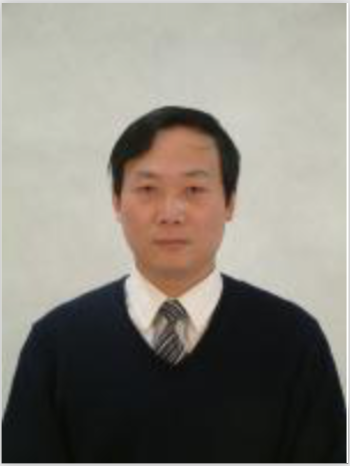}}]{Jiankun Hu}
	receive the Ph.D. degree in control engineering from the Harbin Institute of Technology, China, in 1993, and the master’s degree in computer science and software engineering from Monash University, Australia, in 2000. 
	He was a Research Fellow with Delft University, The Netherlands, from 1997 to 1998, and The University of Melbourne, Australia, from 1998 to 1999. He is a full professor of Cyber Security at the School of Engineering and Information Technology, the University of New South Wales at Canberra, Australia. 
	His main research interest is in the field of cyber security, including biometrics security, where he has published many papers in high-quality conferences and journals including the IEEE TRANSACTIONS ON PATTERN ANALYSIS AND MACHINE INTELLIGENCE. 
	He has served on the editorial boards of up to seven international journals and served as a Security Symposium Chair of the IEEE Flagship Conferences of IEEE ICC and IEEE GLOBECOM. 
	He has obtained nine Australian Research Council (ARC) Grants. He served at the prestigious Panel of Mathematics, Information and Computing Sciences, ARC ERA (The Excellence in Research for Australia) Evaluation Committee 2012.
	
\end{IEEEbiography}
	
\end{document}